\documentclass[letterpaper, 10 pt, conference]{ieeeconf}
\IEEEoverridecommandlockouts
\usepackage[utf8x]{inputenc}
\usepackage{graphicx}
\usepackage{float}
\usepackage[colorinlistoftodos]{todonotes}

\usepackage{atbegshi,picture}
\AtBeginShipoutNext{\AtBeginShipoutUpperLeft{%
  \put(\dimexpr\paperwidth-1cm\relax,-1.5cm){\makebox[0pt][r]{Accepted manuscript IEEE/EMBS Neural Engineering (NER) 2021}}%
}}

\title{\LARGE \bf Inter-subject Deep Transfer Learning for Motor Imagery EEG Decoding}
\author{Xiaoxi Wei$^{1}$, Pablo Ortega$^{1}$ and A. Aldo Faisal$^{1,2}$
\thanks{Brain {\&} Behaviour Lab: $^{1}$ Department of Computing,$^{2}$ Department of Bioengineering, Imperial College London. Correspondence: aldo.faisal@imperial.ac.uk}
}

\begin{document}
\maketitle
\thispagestyle{empty}
\pagestyle{empty}

\begin{abstract}
Convolutional neural networks (CNNs) have become a powerful technique to decode EEG and have become the benchmark for motor imagery EEG Brain-Computer-Interface (BCI) decoding. However, it is still challenging to train CNNs on multiple subjects' EEG without decreasing individual performance. This is known as the negative transfer problem, i.e. learning from dissimilar distributions causes CNNs to misrepresent each of them instead of learning a richer representation. As a result, CNNs cannot directly use multiple subjects' EEG to enhance model performance directly. To address this problem, we extend deep transfer learning techniques to the EEG multi-subject training case. We propose a multi-branch deep transfer network, the Separate-Common-Separate Network (SCSN) based on splitting the network's feature extractors for individual subjects. We also explore the possibility of applying Maximum-mean discrepancy (MMD) to the SCSN (SCSN-MMD) to better align distributions of features from individual feature extractors. The proposed network is evaluated on the BCI Competition IV 2a dataset (BCICIV2a dataset) and our online recorded dataset. Results show that the proposed SCSN (81.8\%, 53.2\%) and SCSN-MMD (81.8\%, 54.8\%) outperformed the benchmark CNN (73.4\%, 48.8\%) on both datasets using multiple subjects. Our proposed networks show the potential to utilise larger multi-subject datasets to train an EEG decoder without being influenced by negative transfer.
\end{abstract}



\vspace{-0.5cm}
\section{Introduction}

Since its first application in BCI in the mid-2010s \cite{walker2015deep}, deep learning has emerged as a powerful technique to learn EEG decoders without biased feature design caused by the limited understanding of human EEG. Convolutional neural networks (CNNs) have outperformed conventional machine learning methods and become the benchmark for motor imagery EEG decoding \cite{3}. However, CNNs are, unlike conventional BCI machine learning \cite{ferrante2015data}, not data-efficient, so combining multiple datasets is essential. EEG features vary across subjects, for example, due to different sensor locations, sensor impedance, and individual brain anatomical and functional organisation. A challenge associated with CNNs is the negative transfer \cite{25}, i.e. learning from dissimilar EEG distributions from different subjects causes CNNs to misrepresent each of them instead of learning a richer representation. 


As a result, unlike other machine learning fields, CNNs cannot directly utilise multiple subjects' EEG to enhance model performance and make up for the data shortage. Additionally, BCI systems usually require a long training period for new users to get a personalised and stabilised decoder, which disrupts the use of BCI in real-world. To address this problem, we combine deep transfer learning with EEG-BCI decoding.

Transfer learning has emerged as a solution to combine data from domains with different distributions \cite{8}. Transfer learning encompasses algorithms which aim to transfer the representations and knowledge from source domains to a target domain. There have been some promising strategies for deep transfer learning in fields like computer vision and natural language processing. For example, fine-tuning is often used to transfer model representations from one task or dataset to another \cite{11}, i.e. networks are pre-trained on one domain and then retrained on another domain. Another strategy is to train a shared network using multiple datasets but split deep network layers for different datasets \cite{12}. In addition, deep domain adaptation \cite{13} is a technique which directly projects source and target distributions into a common distribution with CNNs, and thus data from source domains can benefit the model performance on the target domain.

\begin{figure}[b]
\centering
\includegraphics[width=0.95\columnwidth]{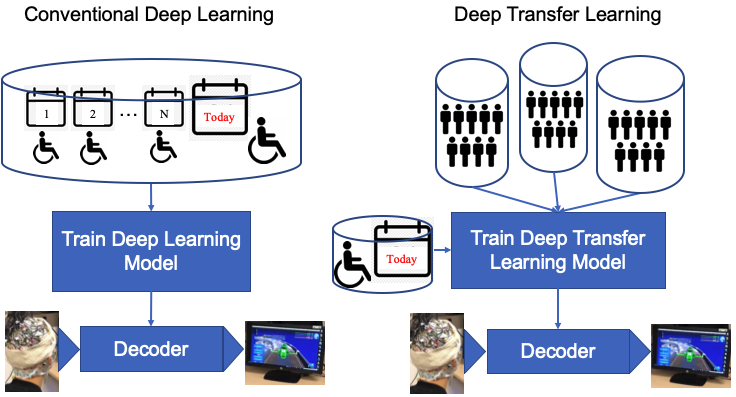} 
\caption{Illustration of Deep Learning and Deep Transfer Learning in BCI. \vspace{-0.5cm}}
\label{fig1}
\end{figure}

Few studies have attempted to adapt the ideas above in EEG decoding. In \cite{14}, a deep transfer network was proposed to transfer the knowledge representation from images to EEG. Following this study, an adversarial network was added into the network to perform better domain adaptation \cite{7}. Another study used a naive idea of pre-training and fine-tuning \cite{15} for motor imagery decoding models. A deep domain adaptation network was used to transfer the data representation from one subject to another \cite{17}, in which deeper layers of the network are separated for the two individuals. However, this domain adaptation network only transfers one subject's distribution to another subject (one-to-one). This limits the amount of data that can be used to train a deep learning model. It is still challenging in the literature for the above deep transfer learning methods to transfer across multiple subjects.

In this study, we designed a multi-subject deep transfer learning network and tested it on the BCI Competition IV 2a Dataset (BCICIV2a dataset) \cite{2} and our own live, online dataset. 

\vspace{-0.25cm}
\section{Methods}

To begin with, we use the benchmark motor imagery decoder \cite{3} as a baseline. The CNN decoder consists of a temporal layer, a spatial layer, a mean pooling layer and a classification layer. The temporal layer extracts time relevant information. The spatial layer performs a spatial convolution across EEG channels. The mean pooling layer averages across features, and finally, the classification layer uses the averaged features to predict a classification label.

\begin{figure}[b]
 \centering
 \includegraphics[width=0.49\textwidth]{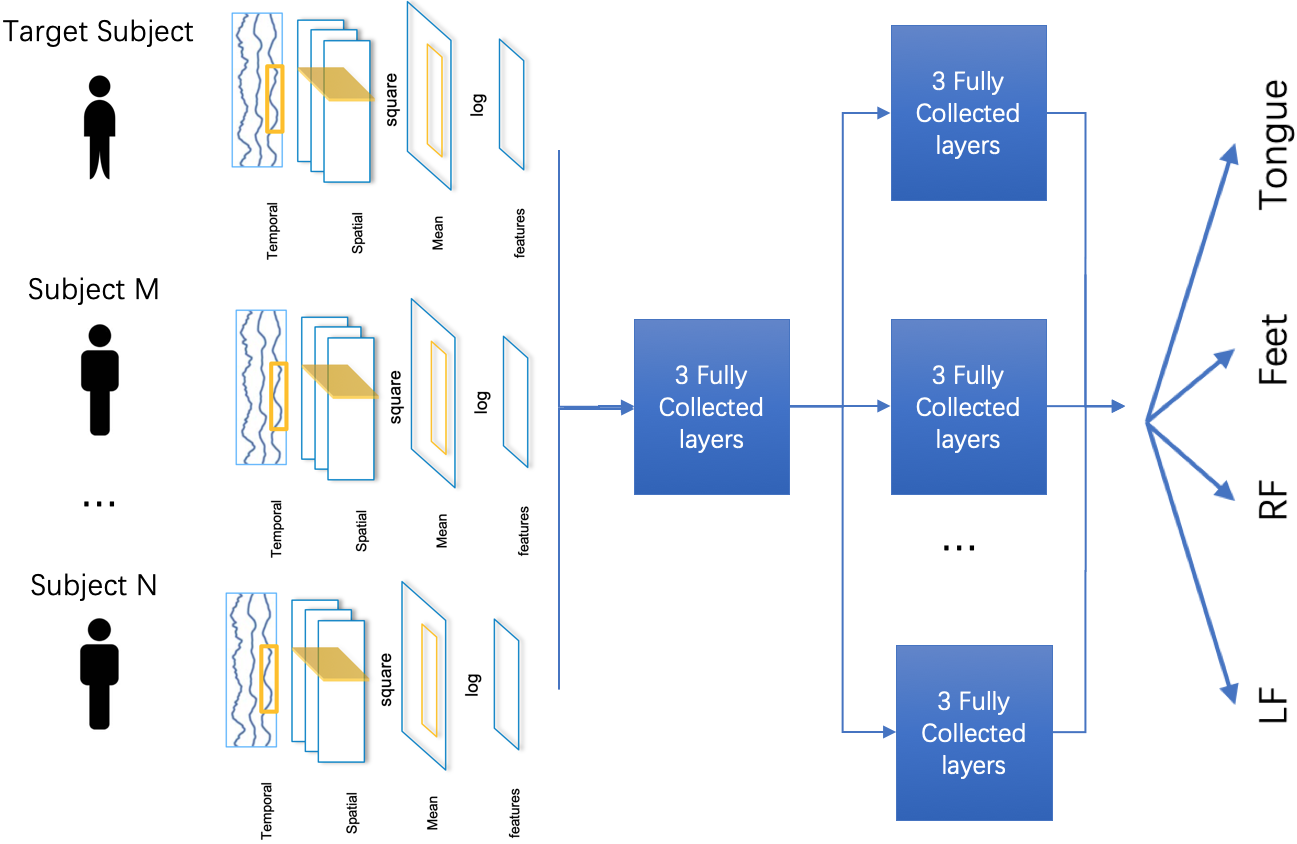} 
 \caption{Architecture of the Separate-Common-Separate Network (SCSN).\vspace{-0.5cm}}
 \label{fig2_d}
\end{figure}

We design a Separate-Common-Separate Network (SCSN) by separating feature extractor of the baseline CNN for individuals, so each of the subjects has their own temporal layer, spatial layer and mean pooling layer, as shown in Figure \ref{fig2_d}. Each network branch extracts subject-specific features, which could avoid negative transfer in the common feature extractor of the baseline CNN. After that, there are three fully connected layers to extract common features of all subjects in the feature space. Unlike the baseline CNN, SCSN extracts common features in the feature space, which could avoid learning from the noise of raw data. Differences in individual's EEG are explained by many aspects, e.g. sensor positions during recording, sensor impedance, and individual brain functionalities. Intuitively, splitting shallow feature extractors could deal with features more related to raw data, i.e. sensor positions and impedance. EEG features which reveal brain functionalities could present in deep layers of the network after several layers of feature extraction. In light of this, the SCSN separate the feature extractors again in deeper layers before the classification layer to handle differences in individual brain functionality.



\begin{equation}
    \mbox{MMD}_i^2(X_{si},X_t) = \\
||\frac{1}{{X_s}_i}\sum_{{x_s}_i\in {X_s}_i}\Phi ({x_s}_i) \nonumber - \frac{1}{X_t}\sum_{x_t\in X_t}\Phi (x_t)||.
\label{Equation1}
\end{equation}


Maximum-mean discrepancy (MMD) \cite{53} is a metric which measures the distance between two datasets in kernel space. For the Separate-Common-Separate Network with MMD (SCSN-MMD), we compute MMD between the target and each source subject in the separate deep feature extractors and add it into the loss function. Equation \ref{Equation1} shows the computation of MMD. According to the literature \cite{19}, we use an RBF kernel with the mean L2 distance of data as the variance. $L_c + \lambda \sum_{i=1}^{N} \mbox{MMD}_i^2(X_{si},X_t)$ shows the loss function, where $L_c$ represents the classification loss, N represents the number of source subjects and $\lambda$ represents the balancing factor between MMD loss and the classification loss. In addition, SCSN has three fully connected layers for individual subjects before the classification layer. We compute MMD for each of the three layers across subjects. The $\mbox{MMD}_i^2$ loss is a weighted average MMD of each of the three layers. Their averaging weights are $\frac{1}{6},\frac{1}{3},\frac{1}{2}$ respectively; this is to increase the significance of deeper layers. We also match samples with the same label when we compute MMD. This ensures the MMD loss represents the distributional distance between data with the same label from different subjects. By projecting the target subject (the new BCI user) and source subjects (supplementary subjects) into a similar domain, the MMD constraint could benefit the model performance of the target subject with more data.

\vspace{-0.25cm}
\section{Evaluation}
\subsection{Dataset} The BCI Competition IV 2a Dataset (BCICIV2a dataset \cite{2}) dataset is frequently used in the literature for motor imagery decoding. The dataset consists of nine subjects. For each subject, two sessions were recorded, and each session consists of 288 four-second trials of four motor imageries (left hand, right hand, both feet and tongue). There are 22 channels covering the scalp. Five of the nine subjects with highest data quality (subject A01, A03, A07, A08, A09) are selected from the dataset according to published results of the competition \cite{2}.

Additionally, we have collected our own online recorded dataset using the Cybathlon 2020 BCI game (https://cybathlon.ethz.ch/en/event/disciplines/bci) with the same hardware set up as our previous work in Cybathlon 2016 \cite{23}. The experimental procedures involving human subjects described in this paper were approved by the Science Engineering Technology Research Ethics Committee of Imperial College London. In the Cybathlon BCI game, subjects attempt to control a virtual wheelchair on a winding race track to go straight, turn left, turn right and turn on the headlight. Our dataset consists of five offline sessions from five right-handed subjects (male), one offline session and one online session from our target subject (right-handed, male), i.e. the pilot. Each session contains 300 5-second trials of four motor imageries (relaxing, left hand, right hand and both feet) of 64 channels covering the whole scalp. In offline sessions, the virtual wheelchair in the Cybathlon BCI game automatically ran on the winding track and subjects were asked to imagine the movements according to actions of the wheelchair. In the online session, the pilot used the baseline CNN model trained on his offline session to perform real-time control to the virtual wheelchair. 

\begin{figure}[b]
\centering
 \vspace{0.2cm}
\includegraphics[width=1.0\linewidth]{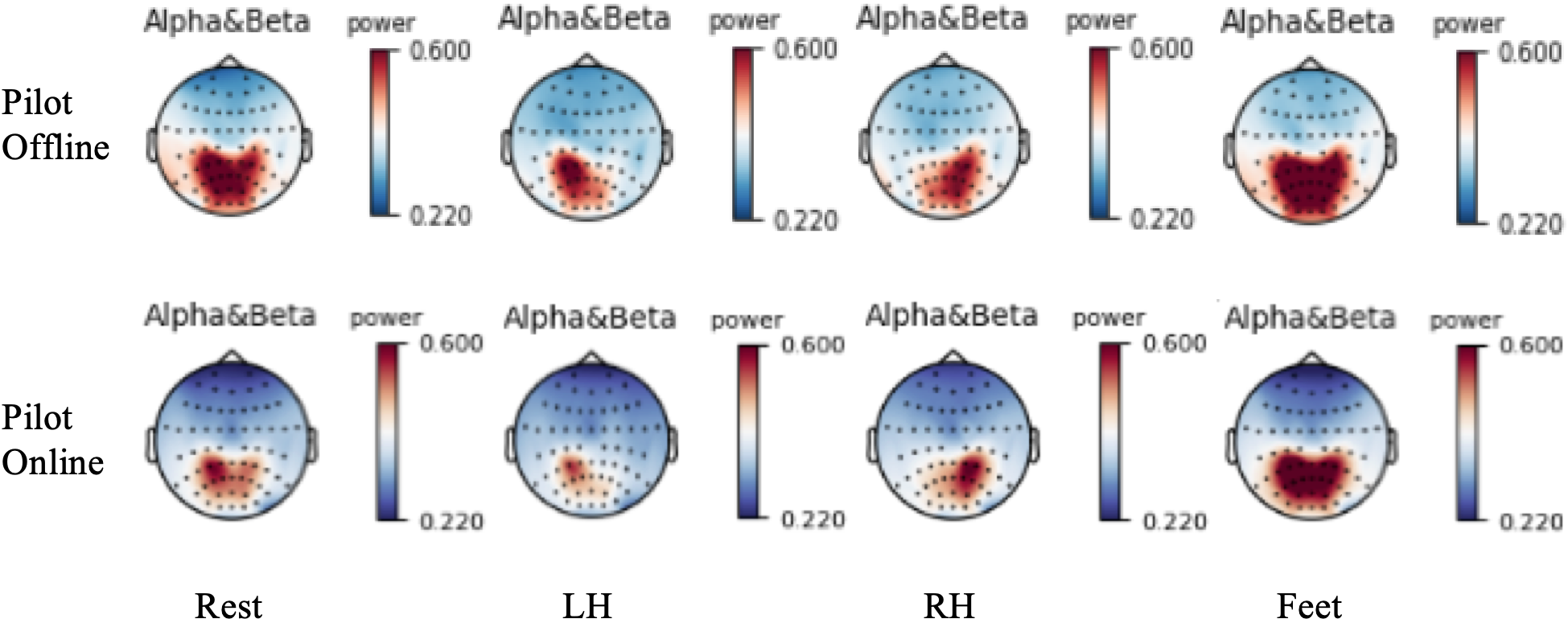} 

\caption{Topographies of the pilot subject in our dataset. }
\label{fig4}
\end{figure}

Figure \ref{fig4} shows topographic EEG maps of the target subject in our dataset. The topographies are computed with the averaged power of each channel across EEG trials. The four columns stand for alpha\&beta band (8Hz - 30Hz) topographic maps (in dB) of the rest, left-hand (LH), right-hand (RH) and feet motor imageries of the offline session and the online pilot session.

\subsection{Data processing pipeline}
Both the BCICIV2a dataset and online dataset use the same preprocessing pipeline. First, we perform a 50Hz notch filter and a bandpass filter between 1Hz and 100Hz. We then crop each trial into 2-second trials with an overlap of 1.9 seconds to better fit the real-time setup. Although using shorter trials may decrease decoding accuracies, it benefits the transition time when subjects attempt to switch from one command to another and the decoding time of the algorithm.

After pre-processing, datasets are split into training sets, validation sets and test sets. For the BCICIV2a dataset, we simulate the calibration period in real-world BCI by including the target subject's first 120 trials of the second session into the training set. The training set consists of the first session of all five subjects as well as the target subject's first 120 trials of the second session, so we train a respective model for each subject in our experiments. The validation set contains the 120th to the 144th trials of the target subject's second session. The last 144 trials in the second session form the test set. Similarly, for our online dataset, the training set of the transfer learning models consists of the offline session of all six subjects as well as the first 100 trials of the online session of the pilot subject. The validation set contains the 100th to the 160th trials. The last 140 trials in the second session form the test set. We select 11 channels covering the motor cortex in our online dataset to reduce online decoding delay. These channels are FC1, FC2, C3, C4, CP5, CP1, CP2, CP6, P3, Pz and P4. 


\vspace{-0.25cm}
\section{Results}

We tested the baseline CNN, SCSN and SCSN-MMD on both BCICIV2a dataset and our online recorded dataset. There are three main observations from the experiment.

\begin{figure}
  \centering
  \vspace{0.2cm}
  \includegraphics[width=.9\linewidth]{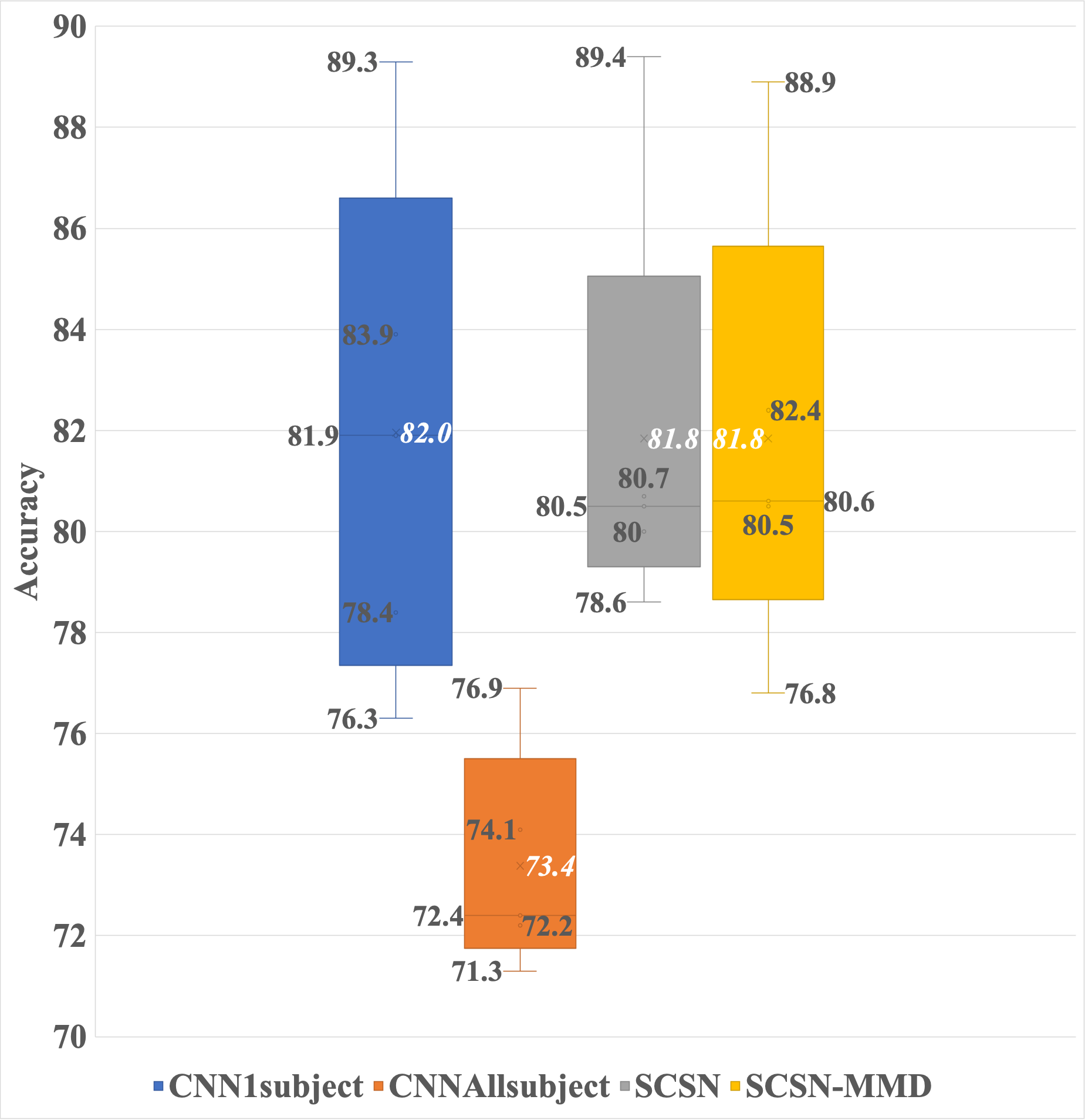}  
\caption{Accuracies of models on BCICIV2a Dataset. The accuracies in black represent the accuracy of each subject, and the accuracies in white show the mean accuracy of different subjects.}
\label{fig6}
\end{figure}

First, the baseline CNN encountered a significant decrease in accuracy when it is trained on multiple subjects. The blue box in Figure \ref{fig6} is the accuracy of baseline CNN trained on the target subject's EEG (session one and the calibration trials of session two). The orange box in Figure \ref{fig6} is the accuracy of baseline CNN trained on the whole training set described in the Evaluation section (Note that each subject has a respective model). On the BCICIV2a dataset, we can observe that the average classification accuracy of the baseline CNN dropped significantly from 82\% to 73.4\% with data from multiple subjects. Similarly, for the online pilot, the accuracy decreased from 54.7\% to 48.8\%. The results support our assumption that the baseline CNN encounters negative transfer.

The main finding of this study is that the multi-branch SCSN and SCSN-MMD outperformed the baseline CNN on the training set with multiple subjects, shown in figure \ref{fig6}. Both SCSN and SCSN-MMD used a batch size of 30 for each branch, so $30N$ trials are forwarded to the network in one training step, where $N$ is the number of subjects. The target subject has more samples than other source subjects because of the additional calibration period, so we randomly duplicate samples, with a balanced size for labels, in each training set of the source subjects. This is to balance the learning procedure of different branches. The average accuracies of the SCSN and SCSN-MMD reached 81.8\%, which are significantly higher than the accuracy of the baseline CNN (73.4\%) using multiple subjects' EEG. We observed similar results on our online recorded dataset. The SCSN (53.2\%) and SCSN-MMD (54.8\%) outperformed the multi-subject baseline (48.8\%). Note that in the testing procedure of SCSN and SCSN-MMD, EEG trials only go through the target branch, so the main difference between the proposed network and the baseline is the networking training.

Finally, comparing SCSN and SCSN-MMD, we did not observe an obvious increase in decoding accuracies by adding MMD constraints to our multi-subject networks. The performance of the SCSN-MMD maintained 81.8\% on the BCICIV2a dataset and increased, slightly, by 1.6\% on the online dataset compared with the SCSN. All experiments are done with an RTX 2080Ti GPU. Approximate training time for the baseline and SCSN on multiple subjects are both 700 to 800 seconds. The training time of SCSN-MMD is around 1000 seconds. Single trial testing time for all networks are within 0.1 second. 

\vspace{-0.25cm}
\section{Discussion}

The SCSN and SCSN-MMD outperformed the baseline on multiple subjects' data (Figure \ref{fig6}). A significant observation from the results is that, for inter-subject deep transfer learning, it is essential to have an individual feature extractor for each subject, and this is the main differences between the proposed networks and the baseline CNN. There are many differences in features of EEG, for example, different channel positions and impedance. We suggest that these inter-subject differences are closely related to raw EEG data and should to be processed in shallow layers of the decoder. Besides, brain functionality of different subjects could be different performing the same motor imagery \cite{52}. This could be revealed in deeper layers of a decoder after feature extraction in shallow layers. Our proposed networks have separate branches for individual subjects in both shallow and deep layers. It considers these differences in individual feature extractors and thus outperform the multi-subject baseline. 

Although the SCSN-MMD also tackle the negative transfer problem, we did not observe significant benefits of applying constraints in our multi-branch networks. The training time of SCSN-MMD is around 20\% longer than SCSN due to additional calculation of MMD. Performance of the SCSN and SCSN-MMD is not significantly different. Compared with \cite{17}, in which MMD is used to transfer from one subject to the other, our study extends the scenario to multiple subjects and compute MMD between the target and each source set, which introduces extra complexity. This may be one reason why applying MMD did not benefit the transfer. To improve the MMD calculation, we could apply multi-kernel methods \cite{13} in future studies, where $\Phi$ in the Equation \ref{Equation1} could be extended to multiple Gaussian kernels instead of one to achieve a better projection of distributions.

One limitation of this study is that, although the proposed networks outperformed the multi-subject baseline CNN, they reach similar accuracy with single-subject baseline. In this study, we used five subjects as the source set, which is relatively a small amount of data for deep transfer learning. This could be one reason why the models did not outperform the single-subject baseline. Larger datasets should be used in future studies to achieve a higher classification accuracy - multi-branch deep transfer learning is ideal for this purpose. Since individual feature extractors can handle different distributions, multi-branch networks like SCSN could be potential solutions to make use of datasets of different subjects with varying sensor sizes, sensor locations and devices. 


\vspace{-0.25cm}
\section{Conclusion}

We have designed a multi-subject deep transfer learning network, the SCSN, and tested it on both BCICIV2a dataset and our online recorded dataset. Results show that our proposed network can tackle the negative transfer problem in the benchmark CNN model, and have outperformed the multi-subject baseline. We believe that our work is an essential conceptual proof towards successfully exploiting multi-subject information in EEG decoding using multi-branch deep transfer learning.

\bibliographystyle{ieeetrans}
\bibliography{bibliography.bib}

\begin{thebibliography}{10}
\providecommand{\url}[1]{#1}
\csname url@samestyle\endcsname
\providecommand{\newblock}{\relax}
\providecommand{\bibinfo}[2]{#2}
\providecommand{\BIBentrySTDinterwordspacing}{\spaceskip=0pt\relax}
\providecommand{\BIBentryALTinterwordstretchfactor}{4}
\providecommand{\BIBentryALTinterwordspacing}{\spaceskip=\fontdimen2\font plus
\BIBentryALTinterwordstretchfactor\fontdimen3\font minus
  \fontdimen4\font\relax}
\providecommand{\BIBforeignlanguage}[2]{{%
\expandafter\ifx\csname l@#1\endcsname\relax
\typeout{** WARNING: IEEEtran.bst: No hyphenation pattern has been}%
\typeout{** loaded for the language `#1'. Using the pattern for}%
\typeout{** the default language instead.}%
\else
\language=\csname l@#1\endcsname
\fi
#2}}
\providecommand{\BIBdecl}{\relax}
\BIBdecl

\bibitem{walker2015deep}
I.~Walker, M.~Deisenroth, and A.~Faisal, ``Deep convolutional neural networks
  for brain computer interface using motor imagery,'' \emph{Imperial College,
  Tech Report}, p.~68, 2015.

\bibitem{3}
R.~T. Schirrmeister, J.~T. Springenberg, L.~D.~J. Fiederer, M.~Glasstetter,
  K.~Eggensperger, M.~Tangermann, F.~Hutter, W.~Burgard, and T.~Ball, ``Deep
  learning with convolutional neural networks for eeg decoding and
  visualization,'' \emph{Human brain mapping}, vol.~38, no.~11, pp. 5391--5420,
  2017.

\bibitem{ferrante2015data}
A.~Ferrante, C.~Gavriel, and A.~Faisal, ``Data-efficient hand motor imagery
  decoding in eeg-bci by using morlet wavelets \& common spatial pattern
  algorithms,'' in \emph{IEEE/EMBS Neural Engineering (NER)}, vol.~7.\hskip 1em
  plus 0.5em minus 0.4em\relax IEEE, 2015, pp. 948--951.

\bibitem{25}
S.~J. Pan and Q.~Yang, ``A survey on transfer learning,'' \emph{IEEE Trans. on
  Knowledge and Data Eng.}, vol.~22, no.~10, pp. 1345--1359, 2009.

\bibitem{8}
V.~Jayaram, M.~Alamgir, Y.~Altun, B.~Scholkopf, and M.~Grosse-Wentrup,
  ``Transfer learning in brain-computer interfaces,'' \emph{IEEE Comp. Intel.
  Magazine}, vol.~11, no.~1, pp. 20--31, 2016.

\bibitem{11}
J.~Yosinski, J.~Clune, Y.~Bengio, and H.~Lipson, ``How transferable are
  features in deep neural networks?'' in \emph{Adv. in neural information
  processing systems (NIPS)}, 2014, pp. 3320--3328.

\bibitem{12}
J.-T. Huang, J.~Li, D.~Yu, L.~Deng, and Y.~Gong, ``Cross-language knowledge
  transfer using multilingual deep neural network with shared hidden layers,''
  in \emph{2013 IEEE Intl. Conf. on Acoustics, Speech and Signal
  Processing}.\hskip 1em plus 0.5em minus 0.4em\relax IEEE, 2013, pp.
  7304--7308.

\bibitem{13}
M.~Long, Y.~Cao, J.~Wang, and M.~I. Jordan, ``Learning transferable features
  with deep adaptation networks,'' \emph{arXiv preprint arXiv:1502.02791},
  2015.

\bibitem{14}
C.~Tan, F.~Sun, and W.~Zhang, ``Deep transfer learning for eeg-based brain
  computer interface,'' in \emph{2018 IEEE Intl. Conf. on Acoustics, Speech and
  Signal Processing (ICASSP)}.\hskip 1em plus 0.5em minus 0.4em\relax IEEE,
  2018, pp. 916--920.

\bibitem{7}
C.~Tan, F.~Sun, B.~Fang, T.~Kong, and W.~Zhang, ``Autoencoder-based transfer
  learning in brain--computer interface for rehabilitation robot,'' \emph{Intl.
  J. Adv. Robotic Sys.}, vol.~16, no.~2, p. 1729881419840860, 2019.

\bibitem{15}
S.~Sakhavi and C.~Guan, ``Convolutional neural network-based transfer learning
  and knowledge distillation using multi-subject data in motor imagery bci,''
  in \emph{8th Intl. IEEE/EMBS Conf. on Neural Engineering (NER)}.\hskip 1em
  plus 0.5em minus 0.4em\relax IEEE, 2017, pp. 588--591.

\bibitem{17}
W.~Hang, W.~Feng, R.~Du, S.~Liang, Y.~Chen, Q.~Wang, and X.~Liu,
  ``Cross-subject eeg signal recognition using deep domain adaptation
  network,'' \emph{IEEE Access}, vol.~7, pp. 128\,273--128\,282, 2019.

\bibitem{2}
M.~Tangermann, K.-R. M{\""u}ller, A.~Aertsen, N.~Birbaumer, C.~Braun,
  C.~Brunner, R.~Leeb, C.~Mehring, K.~J. Miller, G.~Mueller-Putz \emph{et~al.},
  ``Review of the bci competition iv,'' \emph{Frontiers in Neuroscience},
  vol.~6, p.~55, 2012.

\bibitem{53}
A.~Gretton, K.~Borgwardt, M.~Rasch, B.~Sch{\"o}lkopf, and A.~Smola, ``A kernel
  method for the two-sample-problem,'' \emph{Advances in neural information
  processing systems}, vol.~19, pp. 513--520, 2006.

\bibitem{19}
A.~Gretton, K.~M. Borgwardt, M.~J. Rasch, B.~Sch{\""o}lkopf, and A.~Smola, ``A
  kernel two-sample test,'' \emph{J Machine Learning Res.}, vol.~13, no.~1, pp.
  723--773, 2012.

\bibitem{23}
P.~Ortega, C.~Colas, and A.~A. Faisal, ``Compact convolutional neural networks
  for multi-class, personalised, closed-loop eeg-bci,'' in \emph{IEEE
  Biomedical Robotics and Biomechatronics (Biorob)}.\hskip 1em plus 0.5em minus
  0.4em\relax IEEE, 2018, pp. 136--141.

\bibitem{52}
A.~Mierau, W.~Klimesch, and J.~Lefebvre, ``State-dependent alpha peak frequency
  shifts: Experimental evidence, potential mechanisms and functional
  implications,'' \emph{Neuroscience}, vol. 360, pp. 146--154, 2017.

\end{thebibliography}

\end{document}